\documentclass[10pt,english]{IEEEtran}

\usepackage{setspace}
\usepackage{microtype}
\usepackage{graphicx}
\usepackage{subfigure}
\usepackage{booktabs} 

\usepackage{hyperref}

\usepackage{algorithm}
\usepackage{algorithmicx}
\usepackage{lipsum}
\usepackage{algpseudocode}

\usepackage{amsmath}
\usepackage{amssymb}
\usepackage{mathtools}
\usepackage{amsthm}

\usepackage[capitalize,noabbrev]{cleveref}

\theoremstyle{plain}
\newtheorem{thrm}{Theorem}

\newtheorem{lemm}{Lemma}

\theoremstyle{definition}
\newtheorem{deffn}{Definition}
\newtheorem{assump}{Assumption}
\theoremstyle{remark}

\usepackage[textsize=tiny]{todonotes}

\def \R {\mathbb{R}}

\def \ssconstant {\beta}
\def \diff {\mathrm{d}}
\def \statedim {p}
\def \controldim {q}
\def \noisedim {p}
\def \totaldim {d}

\def \policy {\pi}

\def \Qmat {Q}
\def \Rmat {R}
\def \truth {A_\star,B_\star}

\def \auxA {\widetilde{A}}

\def \auxB {\widetilde{B}}

\def \argmin {\arg\min}

\newcommand{\Mnorm}[2]{{\left\vert\kern-0.30ex\left\vert #1 
		\right\vert\kern-0.30ex\right\vert}}
\newcommand{\norm}[2]{{\left\vert\kern-0.30ex\left\vert #1 
		\right\vert\kern-0.30ex\right\vert}}
\newcommand{\eigmax}[1]{\boldsymbol{\lambda}_{\max} \left( #1 \right)}
\newcommand{\eigmin}[1]{\boldsymbol{\lambda}_{\min} \left( #1 \right)}

\newcommand{\PP}[1]{%
	\mathbb{P}{\ifthenelse{ \equal{#1}{} }{}{\left(#1\right)}}
}%

\newcommand{\regret}[2]{\mathrm{\bf Reg} \left(#1\right)}

\newcommand{\RiccSol}[1]{\boldsymbol{P}_{#1}}
\newcommand{\Optgain}[1]{\boldsymbol{K}_{#1}}
\newcommand{\Gainmat}[1]{K_{#1}}

\newcommand{\optavecost}[1]{\overline{\mathcal{J}}^\star }

\newcommand{\optdisccost}[1]{\mathcal{J}_\gamma^\star}

\newcommand{\order}[1]{ \mathcal{O} \left(#1\right)}
\newcommand{\orderlog}[1]{\widetilde{\mathcal{O}} \left(#1\right)}

\newcommand{\Amat}[1]{A_{#1}}
\newcommand{\Bmat}[1]{B_{#1}}
\newcommand{\CLmat}[1]{D_{#1}}
\newcommand{\estpara}[1]{{A}_{#1},{B}_{#1}}
\newcommand{\estA}[1]{{A}_{#1}}
\newcommand{\estB}[1]{{B}_{#1}}

\newcommand{\empiricalcovmat}[1]{V_{#1}}

\newcommand{\state}[1]{x_{#1}}
\newcommand{\optstate}[1]{x^{\star}_{#1}}
\newcommand{\optaction}[1]{u^{\star}_{#1}}
\newcommand{\statetwo}[1]{y_{#1}}
\newcommand{\action}[1]{u_{#1}}
\newcommand{\itointeg}[4]{\int\limits_{#1}^{#2} {#3} \diff {#4}}

\newcommand*{\BM}[1]{
	\mathbb{W}_{\ifthenelse{ \equal{#1}{} }{}{#1}}
}%

\newcommand{\BMcoeff}[1]{C}

\newcommand{\parameter}[1]{A_{#1},B_{#1}}
\newcommand{\normaldist}[2]{\boldsymbol{N} \left( #1, #2 \right)}

\newcommand{\Learnerror}[2]{\Mnorm{\left[#2\right]-\left[\truth\right]}{2}}

\newcommand{\episodetime}[1]{\tau_{#1}}

\newcommand{\regterm}[1]{\alpha_{#1}}
\newcommand{\randommatrix}[1]{\Phi_{#1}}

\newcommand{\trans}[1]{D}

\newcommand{\manifold}[1]{\mathcal{#1}}

\onecolumn

\begin{document}
\title{Regret Analysis of Certainty Equivalence Policies in Continuous-Time Linear-Quadratic Systems}

\author{Mohamad Kazem Shirani Faradonbeh}

\maketitle

\begin{abstract}
	This work theoretically studies a ubiquitous reinforcement learning policy for controlling the canonical model of continuous-time stochastic linear-quadratic systems. We show that randomized certainty equivalent policy addresses the exploration-exploitation dilemma in linear control systems that evolve according to unknown stochastic differential equations and their operating cost is quadratic. More precisely, we establish square-root of time regret bounds, indicating that randomized certainty equivalent policy learns optimal control actions fast from a single state trajectory. Further, linear scaling of the regret with the number of parameters is shown. The presented analysis introduces novel and useful technical approaches, and sheds light on fundamental challenges of continuous-time reinforcement learning.
\end{abstract}

\section{Introduction}
Linear state-space models are of the most popular settings for decision-making in  continuous-time environments. A canonical problem is to minimize quadratic costs subject to state-dynamics that follow stochastic differential equations driven by control actions and Brownian noise. While applications are enormous~\cite{stengel1994optimal,yong1999stochastic,schmidli2007stochastic,lawrence2010learning,oksendal2013stochastic}, little is known about data-driven methods for decision-making under uncertainty. A natural candidate is the randomized certainty equivalent policy that utilizes randomizations together with the Certainty Equivalence principle, and will be the subject of this work.

While the existing literature is rich about reinforcement learning policies for systems following a \emph{discrete-time} dynamics~\cite{abbasi2011regret,ouyang2019posterior,faradonbeh2020adaptive,faradonbeh2020input,cassel2020logarithmic,lale2020logarithmic,ziemann2021uninformative}, 
study of efficient policies for \emph{continuous-time} systems is immature. Early works focus on asymptotic consistency and propose some control policies with linearly growing regrets~\cite{mandl1988consistency,duncan1999adaptive,caines2019stochastic}. Further, offline reinforcement learning algorithms that rely on multiple state trajectories are considered in some settings
~\cite{bian2016adaptive,doya2000reinforcement,wang2020reinforcement,basei2021logarithmic}. 
However, performance analysis of \emph{online} policies that learn from a \emph{single} trajectory of system state to design the control law, are currently sparse~\cite{faradonbeh2022thompson}.  

A fundamental challenge (compared to offline methods) is that an online policy needs to \emph{simultaneously} minimize the cost and estimate the unknown dynamics. The dichotomy of these two contradicting, yet mutually necessary objectives is prevalent in data-driven decision-making. On one hand, we need learning, estimation, and exploration, as apposed to earning, control, and exploitation on the other hand. Importantly, accurate estimation is necessary for good control and for efficiency, while sub-optimal control actions are required in order to have rich data for estimating accurately. 

This work establishes that the popular randomized certainty equivalent reinforcement learning policy balances the trade-off between the exploration and exploitation. We present  Algorithm~\ref{algo1}, which is an episodic randomized certainty equivalent policy for stochastic continuous-time linear systems. We provide its regret analysis indicating efficiency; it learns the optimal control actions fast so that the regret at time $T$ is $\orderlog{{T}^{1/2}}$. Therefore, the per-unit-time sub-optimality gap shrinks with the rate $\orderlog{T^{-1/2}}$ as time proceeds. The presented bound is tight and is obtained under minimal technical assumptions.

To obtain the results, we need to address important challenges. First, analysis of estimation error is needed for sample observations with ill-conditioned information matrices.  
Further, anti-concentration of singular values of random matrices, and full characterization of sub-optimalities in terms of model uncertainties are required. Thus, we develop novel techniques for establishing the rates of identifying the unknown system dynamics matrices based on the data of a single state-input trajectory. Leveraging that together with the effect of diminishing randomizations applied to the parameter estimates, we tightly bound the rates of narrowing down the sub-optimality gap. We also utilize useful results about {Ito integrals} and {random matrices} to precisely capture the additional cost of sub-optimal control actions. En route, different tools from stochastic control, Ito calculus, and stochastic analysis are used, including Hamilton-Jacobi-Bellman equations, Ito Isometry, and martingale convergence theorems~\cite{stengel1994optimal,yong1999stochastic,oksendal2013stochastic}.

The outline of the subsequent sections is provided next. In Section~\ref{ProblemStatement}, we discuss the problem under investigation. 
Section~\ref{AlgoSection} contains the randomized certainty equivalent Algorithm~\ref{algo1}, followed by its theoretical and empirical analyses in Section~\ref{AnalysisSection}.
Technical proofs are provided in the appendices.

\section{Problem Formulation: Continuous-Time Reinforcement Learning} \label{ProblemStatement}
We study reinforcement learning algorithms for an uncertain controlled multidimensional Ito stochastic differential equation~\cite{oksendal2013stochastic}. The state of the plant at time $t$ is denoted by $\state{t} \in \R^{\statedim}$, while the control input is $\action{t} \in \R^{\controldim}$, and we have
\begin{equation} \label{dynamics}
\diff \state{t} = \Amat{\star} \state{t} \diff t + \Bmat{\star} \action{t} \diff t + \BMcoeff{t} \diff \BM{t}.
\end{equation}

In the above dynamics equation of the system state, the stochastic disturbance $\left\{ \BM{t}\right\}_{t \geq 0}$ is a Brownian process. 
Technically, $\BM{t}$ has independent normal increments: for all $0 \leq t_1 \leq t_2 \leq t_3 \leq t_4$, the vectors $\BM{t_2}-\BM{t_1}$ and $\BM{t_4}-\BM{t_3}$ are statistically independent, and  
$$\BM{t_2}-\BM{t_1} \sim \normaldist{0}{\left( t_2-t_1 \right) I_{\noisedim} },$$
where $\normaldist{\cdot}{\cdot}$ is the multivariate normal distribution.
The ${\statedim \times \noisedim}$ matrix $\BMcoeff{}$ might amplify the stochastic disturbance. 

For the plant under consideration, $\Amat{\star}$ is the state evolution matrix and $\Bmat{\star}$ reflects the influence if the control signal. While $\Amat{\star},\Bmat{\star},\BMcoeff{}$ are \emph{unknown}, we aim to design efficient algorithms to minimize the quadratic cost function averaged over time;
\begin{equation} \label{CostEq}
\min\limits_{\left\{\action{t}\right\}_{t \geq 0}}
\limsup\limits_{T \to \infty} \frac{1}{T} \itointeg{0}{T}{ \left(\state{t}^{\top} \Qmat \state{t}+ \action{t}^{\top} \Rmat \action{t}\right) }{t},
\end{equation}

where the symmetric positive definite matrices $\Qmat,\Rmat$ have proper dimensions. Above, the minimum is taken over non-anticipating closed-loop reinforcement learning policies, as elaborated below. The policy determines $\action{t}$ according to the information available at the time, which comprise the state observations $\left\{ \state{s} \right\}_{0 \leq s \leq t}$ and the previously taken actions $\left\{ \action{s} \right\}_{0 \leq s < t}$. Importantly, the policy faces the fundamental exploration-exploitation dilemma, because the plant matrices $\truth$ are unknown. The details of this challenge will be discussed in Section~\ref{AlgoSection}. It is standard to focus on the setting that $\Qmat, \Rmat$ are known, the rationale being that the decision-maker is aware of the objective.  We also assume plant stabilizability:
\begin{assump} \label{StabAssump} 
	There is a matrix $\Gainmat{\star} \in \R^{\controldim \times \statedim}$, such that all eigenvalues of ${\Amat{\star}+\Bmat{\star}\Gainmat{\star}}$ have negative real-parts.
\end{assump}
Assumption~\ref{StabAssump} expresses that by applying $\action{t}=\Gainmat{\star}\state{t}$, the system can operate without unbounded growth in the state. Technically, if we apply the above feedback control law, and solve the differential equation in ~\eqref{dynamics}, it holds that 
\begin{equation} \label{stateevol}
\state{t}=e^{\left(\Amat{\star}+\Bmat{\star}\Gainmat{\star}\right)t} \state{0}+ \itointeg{0}{t}{e^{\left(\Amat{\star}+\Bmat{\star}\Gainmat{\star}\right)(t-s)} \BMcoeff{}}{\BM{s}}.
\end{equation}
According to the above equation, if an eigenvalue of ${\Amat{\star}+\Bmat{\star}\Gainmat{\star}}$ has a non-negative real-part, the state $\state{t}$ grows unbounded with $t$. Thus, Assumption~\ref{StabAssump} is required for a well-posed problem. Otherwise, state explosion renders the cost infinite for all policies~\cite{stengel1994optimal,yong1999stochastic}. Note that $\truth$, and so $\Gainmat{\star}$, are unknown.

In the sequel, we examine effects of uncertainties about $\truth$ on the increase in cost compared to its optimal value. The common assessmet criteria in reinforcement learning is to compare the policy under consideration to the optimal control law $\optaction{t}$ that is decided according to $\truth$. Namely, for generic dynamics matrices $\Amat{},\Bmat{}$, define the feedback matrix $\Optgain{\parameter{}}$ based on $\RiccSol{\parameter{}}$, that solves
\begin{equation} \label{ARiccEq}
\Amat{}^{\top} \RiccSol{\parameter{}} + \RiccSol{\parameter{}} \Amat{} - \RiccSol{\parameter{}} \Bmat{} \Rmat^{-1} \Bmat{}^{\top} \RiccSol{\parameter{}} + \Qmat =0.
\end{equation}

\begin{deffn}
	For generic dynamics matrices $\parameter{}$, define $\Optgain{\parameter{}}= - \Rmat^{-1} \Bmat{}^{\top}\RiccSol{\parameter{}}$, where $\RiccSol{\parameter{}}$ satisfies \eqref{ARiccEq}.
\end{deffn}
So, the unique existence of $\RiccSol{\truth}$ and optimality of the following control law are proven in the literature~\cite{stengel1994optimal,yong1999stochastic}:

\begin{equation} \label{OptimalPolicy} 
\optaction{t}=\Optgain{\Amat{\star},\Bmat{\star}} \optstate{t}, ~~~~~~~~~\text{for all }~~ t \geq 0.
\end{equation}
\begin{thrm} \label{OptimalityProof}
	The matrix $\RiccSol{\truth}$ in \eqref{ARiccEq} uniquely exists, and the linear feedback policy in~\eqref{OptimalPolicy} is optimal.
\end{thrm}
To see the intuition of \eqref{ARiccEq} for obtaining the optimal policy in \eqref{OptimalPolicy}, note that the control action $\action{t}$ directly influences the current cost value, and indirectly affects the future costs according to~\eqref{dynamics}. So, the effects of control actions in the future need to be considered for minimizing the cost function in \eqref{CostEq}, and this consideration is performed by $\RiccSol{\truth}$~\cite{stengel1994optimal,yong1999stochastic}. 

Next, we formulate sub-optimalities and increase in cost due to lack of knowledge about the optimal actions $\optaction{t}$. For a reinforcement learning policy, its \emph{regret} is the total increase in the cost by the time. That is, the gap between the cost the adaptive control law incurs and that of the optimal feedback in~\eqref{OptimalPolicy} is integrated over the interval $\left[0,T\right]$:
{\small \begin{equation*} \label{RegretDefEq}
	\regret{T}{\policy} = \itointeg{0}{T}{ \left( \state{t}^{\top} \Qmat \state{t}+ \action{t}^{\top} \Rmat \action{t} - {\optstate{t}}^{\top} \Qmat \optstate{t} - {\optaction{t}}^{\top} \Rmat \optaction{t} \right) }{t}.
	\end{equation*}}

Note that the stochastic state $\state{t}$ and control signal $\action{t}$ make the regret a random variable. We perform worst-case analysis and bound $\regret{T}{\policy}$ in terms of $T,\statedim, \controldim$. 
If the increasing observations of state and action over time will be effectively leveraged, the policy eventually takes near-optimal actions. So, $\regret{T}{\policy}$ is expected to scale sub-linearly with $T$. However, design of efficient policies with $\orderlog{\sqrt{T}}$ regret and proving performance guarantees for them is challenging, as will be discussed shortly. 

\section{Randomized Certainty Equivalent Policy: Algorithm and Intuition} \label{AlgoSection}
Now, we discuss how randomization of control inputs that are designed according to the Certainty Equivalence principle suffices for learning based control of the plant in \eqref{dynamics}. We aim to have computationally fast algorithms with efficient performance guarantees for minimizing the cost function defined in \eqref{CostEq}. That is, low-regret control laws that can deal with uncertainties about $\truth$. 

First, we explain the challenge of balancing exploration (i.e., estimation) versus exploitation (i.e., control). Then, we study a useful method for estimating the system matrices according to the applied control law and the generated state signal. Based on them, the randomized certainty equivalent adaptive control law that randomizes the parameter estimates to balance estimation and control is discussed, as shown in Algorithm~\ref{algo1}. Finally, we provide theoretical and empirical performance analyses for the proposed algorithm. 

In order to have a policy whose regret is not very large, we need $\action{t} \approx \Optgain{\truth} \state{t}$. Furthermore, since $\truth$ are unknown, the control policy needs to estimate them according to the available trajectory by the time, which is $\left\{ \state{s},\action{s} \right\}_{0 \leq s \leq  t}$. However, if it holds that $\action{s} \approx \Optgain{\truth} \state{s}$, then the coordinates $\action{s}$ of the data $\state{s},\action{s}$ cannot significantly contribute to the estimation procedure; roughly speaking, because they are nothing but linear functions of the state coordinates $\state{s}$. Accordingly, accurate estimation of $\truth$ becomes infeasible, defeating the original purpose. Note that $\truth$ need to be precisely estimated for actuating the plant with near-optimal control inputs. The above-mentioned dilemma is an important challenge and indicates the fact that a good adaptive control law must \emph{randomize} the control inputs $\action{s}$, and so unavoidably it \emph{deviates} from the optimal feedback policy $\optaction{s}=\Optgain{\truth} \optstate{s}$.

Next, we derive an estimator for the unknown dynamics matrices $\truth$. Intuitively speaking, a framework similar to linear regression is used to estimate $\estpara{}$ using the observed trajectories of the state and the input signal. To proceed, suppose that we aim to use samples of the trajectory at $\epsilon$-apart discrete time points; $\left\{ \state{k\epsilon},\action{k\epsilon} \right\}_{k=0}^n$. So, for small $\epsilon$, the stochastic differential equation of the system dynamics in \eqref{dynamics} gives
$$\state{(k+1)\epsilon}-\state{k\epsilon} \approx \left( \Amat{\star} \state{k\epsilon} + \Bmat{\star} \action{k\epsilon} \right) \epsilon + \BMcoeff{}\left(\BM{(k+1)\epsilon} - \BM{k\epsilon}\right).$$
Fitting a linear regression, the least-squares estimate is
$$\argmin\limits_{\estpara{}} \sum\limits_{k=0}^{n-1} \norm{\state{(k+1)\epsilon}-\state{k\epsilon} - \left( \estA{} \state{k\epsilon} + \estB{} \action{k\epsilon} \right) \epsilon}{2}^2,$$
which, letting $\statetwo{s}=\left[\state{s}^{\top},\action{s}^{\top}\right]^{\top}$, leads to
$$\left[\estpara{}\right] = \sum\limits_{k=0}^{n-1} \left(\state{(k+1)\epsilon}-\state{k\epsilon}\right) \statetwo{k\epsilon}^{\top} \left( \sum\limits_{k=0}^{n-1} \statetwo{k\epsilon} \statetwo{k\epsilon}^{\top} \epsilon \right)^{-1}.$$
Therefore, letting $\epsilon \to 0$, we get the continuous-time estimator in \eqref{RandomLSE1} that estimates $\truth$ at the end of every episode of Algorithm~\ref{algo1}, as explained below.

To introduce the episodes of the algorithm, we use the sequence $\left\{ \episodetime{n} \right\}_{n=0}^\infty$ that contains the time points at which the adaptive control law renews its estimates of the unknown system matrices. Namely, during the episode $\episodetime{n} \leq t < \episodetime{n+1}$, Algorithm~\ref{algo1} applies the adaptive feedback policy $\action{t}=\Optgain{\estpara{n}}\state{t}$, where $\estpara{n}$ are the above-mentioned estimates for $\truth$. The episode lengths satisfy
\begin{equation} \label{EpisodeLengthBoundEq}
\underline{\ssconstant} \leq \inf\limits_{n \geq 1} \frac{\episodetime{n+1}-\episodetime{n}} {\episodetime{n}} \leq \sup\limits_{n \geq 1} \frac{\episodetime{n+1}-\episodetime{n}} {\episodetime{n}} \leq \overline{\ssconstant},
\end{equation}
for some constants $\underline{\ssconstant}>0 , \overline{\ssconstant}<\infty$. The rationale for freezing the parameter estimates during the episodes is that the learning procedure can be deferred until collecting enough new observations. Clearly, smaller $\underline{\ssconstant},\overline{\ssconstant}$ mean shorter episodes and more frequent updates in parameter estimates, which gives better exploration. Still, the episode lengths $\episodetime{n+1}-\episodetime{n}$ grow large to preclude unnecessary updates. 

Further, to ensure that the policy is sufficiently committed to explore the environment, a random matrix $\randommatrix{n}$ is added to the least-squares estimate, as shown in \eqref{RandomLSE1}, where $\left\{\randommatrix{n}\right\}_{n=0}^{\infty}$ are $\statedim \times \left(\statedim+\controldim\right)$ random matrices, independent of everything else and of each others, and has independent standard Gaussian entries. The randomized certainty Equivalent reinforcement learning policy is provided in Algorithm~\ref{algo1}.

\begin{algorithm}[H]
	\caption{Randomized Certainty Equivalent Policy} \label{algo1}
	\begin{algorithmic}[1]
		\State {Let $\estpara{0}$ be the initial estimates and $\left\{ \episodetime{n} \right\}_{n=1}^\infty$ satisfy \eqref{EpisodeLengthBoundEq} }
		\For{$n=1,2, \cdots$}
		\While{$\episodetime{n-1} \leq t < \episodetime{n}$}
		\State Apply the feedback $\action{t}=\Optgain{\estpara{n-1}} \state{t}$ 
		\EndWhile
		\State Let $\statetwo{s}=\left[\state{s}^{\top},\action{s}^{\top}\right]^{\top}$ and calculate 
		{\begin{equation} \label{RandomLSE1}
			\left[ \estpara{n} \right]= {\left[ \itointeg{0}{\episodetime{n}}{ \statetwo{s}  }{\state{s}^{\top}} \right]^{\top} \left( \itointeg{0}{\episodetime{n}}{ \statetwo{s} \statetwo{s}^\top  }{s} \right)^{-1} } +  \frac{\randommatrix{n}}{\episodetime{n}^{1/4}}.
			\end{equation}}
		\EndFor
	\end{algorithmic}
\end{algorithm}

The coefficients $\episodetime{n}^{-1/4}$ of the sequence of random matrices are employed to serve a two-fold purpose. On one hand, the scaled random matrix $\episodetime{n}^{-1/4}\randommatrix{n}$ is large enough to significantly randomize the estimates and \emph{explore}. At the same time, $\episodetime{n}^{-1/4}\randommatrix{n}$ is sufficiently \emph{small} to prevent significant deviations from the least-squares estimates and from the optimal actions. Otherwise, large randomizations deteriorate the \emph{exploitation}. 

Note that implementation of Algorithm~\ref{algo1} is fast and requires minimal memory, as one needs to only update the integrals in \eqref{RandomLSE1} in an online fashion. 

\section{Performance Analysis: Regret Bound} \label{AnalysisSection}
Next, we establish Theorem~\ref{BoundsThm} that expresses efficiency of Algorithm~\ref{algo1} in the sense that its regret is $\orderlog{\sqrt{T}}$. 

We suppose that when running Algorithm~\ref{algo1}, the system evolves in a stable manner. That can be equivalently stated as follows: in the plane of complex numbers, all eigenvalues of $\Amat{\star}+\Bmat{\star} \Optgain{\estpara{n}}$ belong to the left half-plane, excluding the imaginary axis. This stability can be ensured in different ways. First, it suffices to find an initial stabilizing policy~\cite{caines1992continuous,duncan1999adaptive,caines2019stochastic}. If such initial stabilizer is available, one can apply it and devote a (relatively short) time period to exploration, such that the collected data provide a coarse approximation of the true matrices $\truth$. Then, it is shown that such coarse-grained approximations are sufficient for stabilization~\cite{duncan1999adaptive,faradonbeh2021bayesian}. Otherwise, to find an initial stabilizer, we can employ Bayesian learning algorithms for a short time period to form a posterior belief about $\truth$. Then, it is known that samples from the posterior belief guarantee high probability stabilization~\cite{faradonbeh2021bayesian}. Since the sampling procedure can be repeated, we can assume that Algorithm~\ref{algo1} remains stable. Further details can be found in the references, as well as in the papers on discrete-time stabilization~\cite{abbasi2011regret,faradonbeh2018bfinite,faradonbeh2019randomized,lale2020explore,chen2021black}.

To establish regret bounds for Algorithm~\ref{algo1}, we assume that the Brownian noise influences all state variables: 
\begin{assump}
	The matrix $\BMcoeff{}$ in \eqref{dynamics} is full-rank.
\end{assump}
This assumption is standard to ensure that the optimal actions can be learned over the course of interactions with the environment \cite{levanony2001persistent,subrahmanyam2019identification,caines2019stochastic,jiang2020learning,basei2021logarithmic,faradonbeh2021bayesian}. Intuitively, it indicates that all state variables have significant roles and a smaller subset of them is \emph{insufficient} for capturing the dynamics of the environment. Now, we present a theoretical performance guarantee for the randomized certainty equivalent policy.

\begin{thrm}  \label{BoundsThm}
	The adaptive control law in Algorithm~\ref{algo1} gives
	\begin{equation*}
	\regret{T}{\policy} = \order{ \totaldim^2 \sqrt{T} \log T},
	\end{equation*}
	where $\totaldim=\statedim+\controldim$ is the total dimension of the system.
\end{thrm}
Above, note that $\totaldim^2$ scales linearly with the number of unknown parameters in $\truth$, which is $\statedim (\statedim+\controldim)$.
\begin{figure}[t!] 
	\centering
	\scalebox{.24}
	{\includegraphics{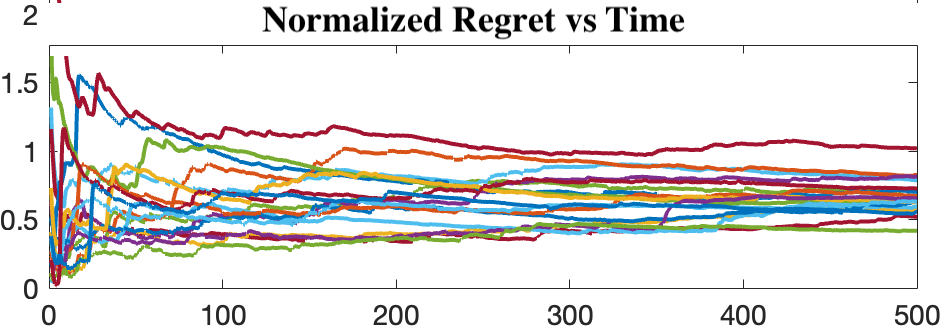}}  
	\caption{The above graph presents curves of the normalized regret $T^{-1/2}\regret{T}{\policy}$ vs $T$, for Algorithm~\ref{algo1}. Multiple replicates of the system are simulated, all of them corroborate Theorem~\ref{BoundsThm} that the normalized regret remains bounded as time grows.}
	\label{RegretFig}
\end{figure}
We experiment Algorithm~\ref{algo1} for adaptive control of an airplane~\cite{bosworth1992linearized}. It is known that the lateral-directional state-space model is of dimensions $\statedim=4$, $\controldim=2$: 
{\small \begin{eqnarray*}
		\Amat{\star} &=& \begin{bmatrix}
			-0.185 & 0.1475 & -0.9825 &  0.1120 \\
			-0.347 & -1.710 & 0.9029 &   -0.58 \times 10^{-6}\\
			1.174 &  -0.0825   & -0.1826 &  -0.44 \times 10^{-7}\\
			0.0 &    1.0 &    0.1429 &   0.0
		\end{bmatrix}, \\
		\Bmat{\star} &=& \begin{bmatrix}
			-0.4470 \times 10^{-3} &   0.4020 \times 10^{-3}\\
			0.3715        &   0.0549 \\
			0.0265        &   -0.0135 \\
			0.0           &   0.0 
		\end{bmatrix}.
\end{eqnarray*}}
Moreover, we let $\BMcoeff{}=0.2 \times I_4$, $\Qmat=I_{\statedim}$, $\Rmat= 0.1 \times I_{\controldim}$, $\episodetime{n}=25 \times 1.2^{n}$, and run Algorithm~\ref{algo1} for the system in \eqref{dynamics}, where $\truth$ are unknown to the algorithm.

Figure~\ref{RegretFig} depicts the normalized regret versus time for Algorithm~\ref{algo1}. The horizontal axis is $T$, while the vertical one corresponds $T^{-1/2} \regret{T}{\policy}$. It shows the result of Theorem~\ref{BoundsThm} that the normalized regret is almost bounded. 

\section{Concluding Remarks and Future Work} \label{ConclusionSection}
This work studies the randomized certainty equivalent reinforcement learning policy in continuous-time stochastic linear systems with quadratic operating cost functions. We presented theoretical performance analysis of the randomized certainty equivalent algorithm showing that it is efficient. More precisely, we established a regret bound that its growing rate as time proceeds is square-root. Further, dependence on the problem dimension is quadratic indicating that the regret grows as a linear function of the number of system parameters.

The presented results motivate interesting directions in the study of reinforcement learning algorithms for continuous-time environments. Finding regret bounds that hold uniformly over time, and deriving performance guarantees in high-dimensional systems with sparse or low-rank dynamics matrices, are interesting problems for future work. Moreover, extension of the presented analysis to reinforcement learning policies under imperfectly observed state, and control laws for nonlinear systems, can be listed as problems of interest for future investigations.

\appendices
\section*{Outline of the Appendices}
In the first appendix, we prove Theorem~\ref{BoundsThm}. Then, the auxiliary lemmas used in the proof are provided.
\section{Proof of Theorem~\ref{BoundsThm}}
Let $\statetwo{s}=\left[\state{s}^{\top},\action{s}^{\top}\right]^{\top}$ and 
$\empiricalcovmat{n} = \itointeg{0}{\episodetime{n}}{\statetwo{s} \statetwo{s}^{\top}}{s}.
$
Replace for $\diff \state{t}$ from \eqref{dynamics} to obtain
\begin{equation*}
\itointeg{0}{\episodetime{n}}{ \statetwo{s} }{\state{s}^{\top}}  =  \itointeg{0}{\episodetime{n}}{ \statetwo{s}  \statetwo{s}^{\top} \left[\truth\right]^{\top}}{s} + \itointeg{0}{\episodetime{n}}{ \statetwo{s} }{\BM{s}^{\top}}  \BMcoeff{}^{\top}.
\end{equation*}
So, we have
{\small \begin{equation*} \label{EstErrorEq}
	\left[ \itointeg{0}{\episodetime{n}}{ \statetwo{s} }{\state{s}^{\top}} \right]^{\top} \empiricalcovmat{n}^{-1} = \left[\truth\right] + \left[ \empiricalcovmat{n}^{-1} \itointeg{0}{\episodetime{n}}{ \statetwo{s} }{\BM{s}^{\top}}  \BMcoeff{}^{\top} \right]^{\top}.
	\end{equation*}}
The above, the triangle inequality, and \eqref{RandomLSE1}, yield to
{\small \begin{equation*}
	\Learnerror{}{\estpara{n}} \leq \Mnorm{\empiricalcovmat{n}^{-1} \itointeg{0}{\episodetime{n}}{ \statetwo{s} }{\BM{s}^{\top}}  \BMcoeff{}^{\top}}{2} + \Mnorm{\episodetime{n}^{-1/4}\randommatrix{n}}{2}.
	\end{equation*}}
Since entries of $\episodetime{n}^{-1/4}\randommatrix{n}$ have  $\normaldist{0}{{\episodetime{n}}^{-1/2}}$ distribution, for $\ssconstant >0$ we have 
{\small $$\log \PP{\Mnorm{\episodetime{n}^{-1/4}\randommatrix{n}}{2} \geq \statedim^{1/2} \left(\statedim+\controldim\right)^{1/2} {\episodetime{n}}^{-1/4} \ssconstant} = \order{{-\ssconstant^2}}.$$}
This, by Borel-Cantelli Lemma and $\episodetime{n} \to \infty$, gives $$\Mnorm{\episodetime{n}^{-1/4}\randommatrix{n}}{2}=\order{\totaldim {\episodetime{n}}^{-1/4} \log^{1/2} \episodetime{n} }.$$

Thus, by Lemma~\ref{SelfNormalizedLem}, $\Learnerror{}{\estpara{n}}$ is at most
\begin{equation}\label{BoundThmProofEq1}
\order{ \totaldim \left(\frac{\log \eigmax{\empiricalcovmat{n}}}{\eigmin{\empiricalcovmat{n}}}\right)^{1/2} + \totaldim \episodetime{n}^{-1/4} \log^{1/2} \episodetime{n}}.
\end{equation}
Now, Lemma~\ref{EmpCovLemma} provides ${\log \eigmax{\empiricalcovmat{n}}} =  \order{\log \episodetime{n}}$. On the other hand, we will establish in the sequel that: 
\begin{equation} \label{BoundThmProofEq2}
\liminf\limits_{n \to \infty} \episodetime{n}^{-1/2} \eigmin{\empiricalcovmat{n}} > 0. 
\end{equation}
Thus, \eqref{BoundThmProofEq1} and \eqref{BoundThmProofEq2} lead to
\begin{equation} \label{EstErrorBoundEq}
\Learnerror{}{\estpara{n}} = \order{ \totaldim \episodetime{n}^{-1/4} \log^{1/2} \episodetime{n} }.
\end{equation}
Next, Lemma~\ref{LipschitzLemma} implies that 
\begin{equation*}
\Mnorm{\Optgain{\estpara{n}} - \Optgain{\truth}}{2}^2 = \order{ \totaldim^2 \episodetime{n}^{-1/2} \log \episodetime{n} }.
\end{equation*} 
Since during the episode $\episodetime{n-1} \leq t < \episodetime{n}$ the parameter estimates are not updated; $\Gainmat{t}=\Optgain{\estpara{n-1}}$, by Lemma~\ref{EmpCovLemma}, we have 
\begin{eqnarray*}
	&&\itointeg{0}{\episodetime{n}}{ \norm{ \left( \Gainmat{t} - \Optgain{\parameter{\star}} \right) \state{t} }{}^2 }{t} \\
	&=& \order{\sum\limits_{k=1}^{n} \frac{\episodetime{k} - \episodetime{k-1}}{\episodetime{k-1}^{1/2}} \totaldim^2  \log \episodetime{k-1} } \\
	&=& \order{\sum\limits_{k=1}^{n} \left( \episodetime{k}^{1/2} - \episodetime{k-1}^{1/2} \right) \totaldim^2  \log \episodetime{n} } \\
	&=& \order{ \episodetime{n}^{1/2} \totaldim^2  \log \episodetime{n} },
\end{eqnarray*}
where in the last two equalities above we used \eqref{EpisodeLengthBoundEq}.

Moreover, considering the matrix $\Amat{\star}+\Bmat{\star}\Optgain{\truth}$, since all of its eigenvalues are negative-real, the matrix denoted by $E_t$ in Lemma~\ref{GeneralRegretThm} decays exponentially with $t$. So, we have 
$${\itointeg{0}{T}{ \left( \state{t}^{\top} E_{T-t} \left( \Gainmat{t}-\Optgain{\truth} \right) \state{t} \right) }{t}} = \order{\log^2 T}.$$ 

Thus, plugging the above two results in Lemma~\ref{GeneralRegretThm}, it implies that
the statement of Theorem~\ref{BoundsThm} holds.

As such, to finish this proof, we can prove \eqref{BoundThmProofEq2}. To do so, note that \eqref{EpisodeLengthBoundEq} together with Lemma~\ref{EmpCovLemma} give
\begin{equation} \label{BoundThmProofEq3}
\liminf\limits_{k \to \infty} \episodetime{k}^{-1} \eigmin{ \itointeg{\episodetime{k-1}}{\episodetime{k}}{ \state{t}\state{t}^{\top} }{t}} >0.
\end{equation}

Thus, for getting \eqref{BoundThmProofEq2}, it is enough to establish 
\begin{equation} \label{BoundThmProofEq4}
\liminf\limits_{n \to \infty} \eigmin{\sum\limits_{k=\ell}^{n-1} \frac{\episodetime{ k}}{\episodetime{n}^{1/2}} \begin{bmatrix}
	I_{\statedim} \\ \Optgain{\estpara{k}}
	\end{bmatrix} \begin{bmatrix}
	I_{\statedim} \\ \Optgain{\estpara{k}}
	\end{bmatrix}^{\top}   } >0,
\end{equation}
for some $0 \leq \ell <n-1$.

For $\epsilon>0$, consider the case of the above-mentioned least eigenvalue being strictly smaller than $\epsilon$. Further, denote by $\manifold{M}_n(\epsilon)$, a set that contains all matrices such as $\left[\estpara{k}\right]_{k=\ell}^{n-1}$, for which the upper bound $\epsilon$ on the least eigenvalue in \eqref{BoundThmProofEq4} occurs. More precisely, define the ${(\statedim+\controldim) \times \statedim (n-\ell)}$ matrix $P_{\ell,n}$ as
\begin{equation*}
\left[ \episodetime{\ell}^{1/2} \episodetime{n}^{-1/4} \begin{bmatrix}
I_{\statedim} \\
\Optgain{\estpara{\ell}}
\end{bmatrix} , \cdots, \episodetime{n-1}^{1/2} \episodetime{n}^{-1/4} \begin{bmatrix}
I_{\statedim} \\
\Optgain{\estpara{n-1}}
\end{bmatrix}\right].
\end{equation*} 
Then, let
{\small \begin{equation*}
	\manifold{M}_n(\epsilon) = \left\{ \left[\estpara{\ell}, \cdots, \estpara{n-1}\right] : \eigmin{P_{\ell,n} P_{\ell,n}^{\top}} \leq \epsilon \right\}.
	\end{equation*}}
Now, note that the set of all matrices 
\begin{equation*}
F_n = \begin{bmatrix}
\episodetime{\ell}^{1/2} \episodetime{n}^{-1/4} I_{\statedim}  & \cdots & \episodetime{n-1}^{1/2} \episodetime{n}^{-1/4} I_{\statedim} \\
\episodetime{\ell}^{1/2} \episodetime{n}^{-1/4} \Gainmat{\ell} &  \cdots & \episodetime{n-1}^{1/2} \episodetime{n}^{-1/4} \Gainmat{n-1}
\end{bmatrix},
\end{equation*}
for which there is $v \in \R^{\statedim+\controldim}$ satisfying $\norm{v}{2}=1 \text{, }F_n^{\top}v=0,$ 
is of dimension $\statedim+\controldim-1+(n-\ell)(\controldim-1)$, as follows:
\begin{enumerate}
	\item 
	The set of unit $\statedim+\controldim$ dimensional vectors is (a sphere) of dimension $\statedim+\controldim-1$.
	\item 
	Write $v=\left[v_1^{\top},v_2^{\top}\right]^{\top}$, for $v_1 \in \R^{\statedim}$ and $v_2 \in \R^{\controldim}$. So, $F_n^{\top}v=0$, if and only if $\Gainmat{k}^{\top}v_2=-v_1$, for all $k=\ell,\cdots, n-1$. This means every column of $\Gainmat{k}$ is in a certain hyperplane in $\R^{\controldim}$. 
\end{enumerate}  

According to Lemma~\ref{OptManifoldLemma}, the dimension of $\manifold{M}_n(0)$ is at most 
$$\statedim + \controldim - 1+ (\controldim-1) (n-\ell)+ (n-\ell) \statedim^2,$$
and it lives in a $\statedim (\statedim+\controldim)(n-\ell)$ dimensional space. So, the difference between the dimensions is
$$m=\left(\statedim \controldim - \controldim+1\right) (n-\ell) - \statedim-\controldim+1.$$

Further, if $\ell$ is sufficiently large so that $\episodetime{\ell}^{-1} \episodetime{n}^{1/2} \epsilon <1$, then for every $\left[\estpara{k}\right]_{k=\ell}^{n-1} \in \manifold{M}_n(\epsilon)$, there exists some $\left[\auxA_k,\auxB_k\right]_{k=\ell}^{n-1} \in \manifold{M}_n(0)$, such that 
\begin{equation*}
\max\limits_{\ell \leq k \leq n-1} \Mnorm{\left[\estpara{k}\right] - \left[\auxA_k,\auxB_k\right]}{2} = \order{\episodetime{k}^{-1/2} \episodetime{n}^{1/4} \epsilon^{1/2}}.
\end{equation*}

Next, we use the above result to bound the probability of $\manifold{M}_n(\epsilon)$. Note that the random matrices $\left\{\randommatrix{k}\right\}_{k=0}^{n-1}$ are independent, and entries of $\episodetime{k}^{-1/4}\randommatrix{k}$ are independent identically distributed $\normaldist{0}{\episodetime{k}^{-1/2}}$ random variables. Recall that the difference between the dimensions of $\manifold{M}_n(0)$ and the space it is in, is $m$, as defined above. Hence, since $\episodetime{\ell} \leq \episodetime{k}$, we have
\begin{equation*}
\PP{\manifold{M}_n(\epsilon)} = \left[ \order{\episodetime{\ell}^{1/4} \episodetime{\ell}^{-1/2} \episodetime{n}^{1/4}  \epsilon^{1/2}} \wedge 1 \right]^{m}.
\end{equation*}
Letting $\ell = n-5$, we have $m\geq 5$. Further, if $\epsilon$ is small enough to satisfy 
$ \order{ \episodetime{n}^{1/4} \episodetime{\ell}^{-1/4} \epsilon^{1/2} } < 1 ,$
we have 
$$\sum\limits_{n=5}^{\infty} \PP{\manifold{M}_n(\epsilon)}  
<  \infty.$$
The above, by Borel-Cantelli Lemma, implies \eqref{BoundThmProofEq4}, which completes the proof. 

\section{Auxiliary Results}
This appendix presents the lemmas used in the proof of Theorem~\ref{BoundsThm} in the previous appendix. First, Lemma~\ref{GeneralRegretThm} expresses the regret in terms of the deviations of the control input from the optimal feedback control law. Then, in Lemma~\ref{SelfNormalizedLem} we present the growth rates of stochastic integrals normalized by the empirical state-input covariance matrix. Lemma~\ref{LipschitzLemma} establishes that $\Optgain{\estpara{}}$ is a (Lipschitz) continuous function of $\estpara{}$. Next, the state empirical covariance matrix is shown to converge to a positive definite limit in Lemma~\ref{EmpCovLemma}. Finally, the result of Lemma~\ref{OptManifoldLemma} provides the dimension of the optimality manifold. Proofs are omitted here due to limited space, but they are available in~\cite{faradonbeh2021efficient}.

\begin{lemm} \label{GeneralRegretThm}
	Suppose that $\Gainmat{t} $ is piecewise continuous and $\action{t}=\Gainmat{t}\state{t}$ is applied to the system in \eqref{dynamics}. Define 
	$\Optgain{\star}=\Optgain{\truth}$, $\CLmat{\star}=\Amat{\star}+\Bmat{\star}\Optgain{\star}$, $E_t=e^{\CLmat{\star}^{\top}t} \RiccSol{\truth} e^{\CLmat{\star}t} \Bmat{\star}$, 
	{\small \begin{eqnarray*}
			\regterm{T} = \itointeg{0}{T}{ \left( \norm{ \left( \Gainmat{t} - \Optgain{\star} \right) \state{t} }{}^2 - 2 \state{t}^{\top} E_{T-t} \left( \Gainmat{t}-\Optgain{\star} \right) \state{t}\right) }{t}.
	\end{eqnarray*}}
	Then, we have $\regret{T}{\policy}=\order{\regterm{T}}$.
\end{lemm}

\begin{lemm} \label{SelfNormalizedLem}
	For $\statetwo{t}=\left[\state{t}^\top, \action{t}^{\top}\right]^\top$, let $\empiricalcovmat{t}=\itointeg{0}{t}{ \statetwo{s} \statetwo{s}^{\top} }{s}$. Then, it holds that
	$$\Mnorm{ \left(I+ \empiricalcovmat{t} \right)^{-1/2} \itointeg{0}{t}{ \statetwo{s} }{\BM{s}^{\top}} }{2}^2 = \order{ \totaldim^2 \log \eigmax{\empiricalcovmat{t}} }.$$
\end{lemm}

\begin{lemm} \label{LipschitzLemma}
	There exists $\ssconstant_\star < \infty$, such that 
	\begin{equation*}
	\Mnorm{\Optgain{\estpara{}} - \Optgain{\truth} }{} \leq \ssconstant_\star \Learnerror{}{\estpara{}}.
	\end{equation*}
\end{lemm}

\begin{lemm} \label{EmpCovLemma}
	In Algorithm~\ref{algo1}, suppose that all eigenvalues of $\Amat{\star}+\Bmat{\star}\Optgain{\estpara{n}}$ have negative real-parts. Then, the following matrix is deterministic and positive definite:
	\begin{eqnarray*}
		\lim\limits_{n \to \infty} \frac{1}{\episodetime{n+1}-\episodetime{n}} \itointeg{\episodetime{n}}{\episodetime{n+1}}{\state{t}\state{t}^{\top}}{t}. 
	\end{eqnarray*}
\end{lemm}

\begin{lemm} \label{OptManifoldLemma}
	For some fixed $\estpara{0}$, consider
	\begin{eqnarray*}
		\manifold{N}_0 = \left\{ \left[ \estpara{} \right] \in \R^{\statedim \times \left(\statedim + \controldim\right)} : \Optgain{\estpara{}}=\Optgain{\estpara{0}} \right\}.
	\end{eqnarray*} 
	The set $\manifold{N}_0$ is a manifold of dimension $\statedim^2$.
\end{lemm}

\end{document}